\documentclass[10pt,twocolumn,letterpaper]{article}

\usepackage{iccv}
\usepackage{times}
\usepackage{epsfig}
\usepackage{graphicx}
\usepackage{amsmath}
\usepackage{amssymb}
\usepackage[ruled]{algorithm2e} 
\usepackage{bm}
\usepackage{multirow}
\usepackage{subfig}
\usepackage{float}
\usepackage{mathrsfs}
\usepackage{array}
\usepackage{esvect}

\usepackage[pagebackref=true,breaklinks=true,letterpaper=true,colorlinks,bookmarks=false]{hyperref}

\iccvfinalcopy

\ificcvfinal\fi
\begin{document}
\title{ SVDNet for Pedestrian Retrieval}

\author{Yifan Sun$^{\dag}$,\ Liang Zheng$^{\ddag}$,\ Weijian Deng$^{\S}$,\ Shengjin Wang$^{\dag}\thanks{Corresponding Author}$ \\
 $^{\dag}$Tsinghua University \quad $^{\ddag}$University of Technology Sydney \\$^{\S} $University of Chinese Academy of Sciences\\
{\tt\small sunyf15@mails.tsinghua.edu.cn, \{liangzheng06, dengwj16\}@gmail.com, wgsgj@tsinghua.edu.cn }
 }

\maketitle
\thispagestyle{empty}
\begin{abstract}
    This paper proposes the SVDNet for retrieval problems, with focus on the application of person re-identification (re-ID). We view each weight vector within a fully connected (FC) layer in a convolutional neuron network (CNN) as a projection basis. It is observed that the weight vectors are usually highly correlated. This problem leads to correlations among entries of the FC descriptor, and compromises the retrieval performance based on the Euclidean distance. To address the problem, this paper proposes to optimize the deep representation learning process with Singular Vector Decomposition (SVD). Specifically, with the restraint and relaxation iteration (RRI) training scheme, we are able to iteratively integrate the orthogonality constraint in CNN training, yielding the so-called SVDNet. We conduct experiments on the Market-1501, CUHK03, and DukeMTMC-reID datasets, and show that RRI effectively reduces the correlation among the projection vectors, produces more discriminative FC descriptors, and significantly improves the re-ID accuracy. On the Market-1501 dataset, for instance, rank-1 accuracy is improved from 55.3\% to 80.5\% for CaffeNet, and from 73.8\% to 82.3\% for ResNet-50.
\end{abstract}

\section{Introduction}
This paper considers the problem of pedestrian retrieval, also called person re-identification (re-ID). This task aims at retrieving images containing the same person to the query.

\begin{figure}[t]
\begin{center}
\includegraphics[width=1.0\linewidth]{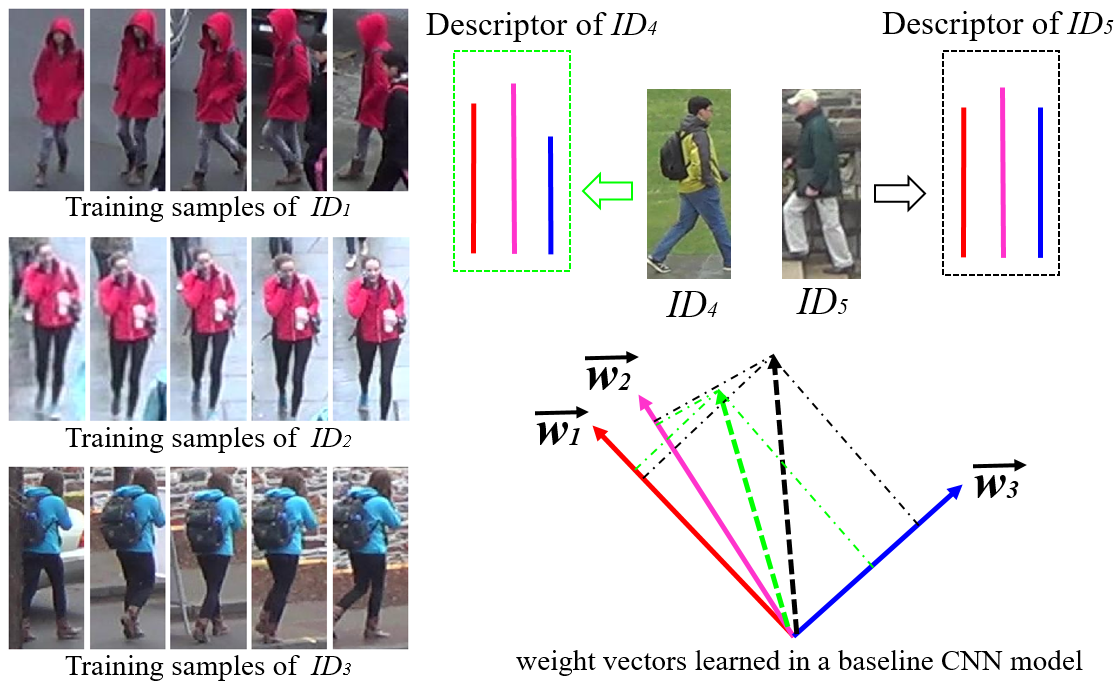}
\end{center}
   \caption{A cartoon illustration of the correlation among weight vectors and its negative effect. The weight vectors are contained in the last fully connected layer, \eg, FC8 layer of CaffeNet \cite{DBLP:conf/nips/KrizhevskySH12} or  FC layer of ResNet-50 \cite{DBLP:conf/cvpr/HeZRS16}. There are three training IDs in red, pink and blue clothes from the DukeMTMC-reID dataset \cite{ristani2016performance}. The dotted green and black vectors denote feature vectors of two testing samples before the last FC layer. Under the baseline setting, the red and the pink weight vectors are highly correlated and introduce redundancy to the descriptors.}
\label{fig:motivation}
\end{figure}
Person re-ID is different from image classification in that the training and testing sets contain entirely different classes. So a popular deep learning method for re-ID consists of 1) training a classification deep model on the training set, 2) extracting image descriptors using the fully-connected (FC) layer for the query and gallery images, and 3) computing similarities based on Euclidean distance before returning the sorted list \cite{DBLP:journals/corr/ZhengYH16,DBLP:journals/corr/ZhengHLY17,DBLP:conf/cvpr/XiaoLOW16,geng2016deep}. 

Our work is motivated by the observation that after training a convolutional neural network (CNN) for classification, the weight vectors within a fully-connected layer (FC) are usually highly correlated. This problem can be attributed to two major reasons. The first reason is related to the non-uniform distribution of training samples. This problem is especially obvious when focusing on the last FC layer. The output of each neuron in the last FC layer represents the similarity between the input image and a corresponding identity. After training, neurons corresponding to similar persons (\ie, the persons who wear red and pink clothes) learns highly correlated weight vectors, as shown in Fig. \ref{fig:motivation}. The second is that during the training of CNN, there exists few, if any, constraints for learning orthogonalization. Thus the learned weight vectors may be naturally correlated.

Correlation among weight vectors of the FC layer compromises the descriptor significantly when we consider the retrieval task under the Euclidean distance. In fact, a critical assumption of using Euclidean distance (or equivalently the cosine distance after $\ell_2$-normalization) for retrieval is that the entries in the feature vector should be possibly independent. However, when the weight vectors are correlated, the FC descriptor -- the projection on these weight vectors of the output of a previous CNN layer --  will have correlated entries. This might finally lead to some entries of the descriptor dominating the Euclidean distance, and cause poor ranking results. For example, during testing, the images of two different persons are passed through the network to generate the green and black dotted feature vectors and then projected onto the red, pink and blue weight vectors to form the descriptors, as shown in Fig. \ref{fig:motivation}. The projection values on both red and pink vectors are close, making the two descriptors appear similar despite of the difference projected on the blue vector. As a consequence, it is of vital importance to reduce the redundancy in the FC descriptor to make it work under the Euclidean distance.

To address the correlation problem, we proposes SVDNet, which is featured by an FC layer containing decorrelated weight vectors. We also introduce a novel three-step training scheme. In the first step, the weight matrix undergoes the singular vector decomposition (SVD) and is replaced by the product of the left unitary matrix and the singular value matrix. Second, we keep the orthogonalized weight matrix fixed and only fine-tune the remaining layers. Third, the weight matrix is unfixed and the network is trained for overall optimization. The three steps are iterated to approximate orthogonality on the weight matrix. Experimental results on three large-scale re-ID datasets demonstrate significant improvement over the baseline network, and our results are on par with the state of the art.

\section{Related Work}

\textbf{Deep learning for person re-ID.} In person re-ID task, deep learning methods can be classified into two classes: similarity learning and representation learning. The former is also called deep metric learning, in which image pairs or triplets are used as input to the network \cite{DBLP:conf/eccv/VariorSLXW16,DBLP:conf/eccv/VariorHW16,DBLP:conf/cvpr/AhmedJM15,DBLP:conf/cvpr/LiZXW14,cheng2016person,DBLP:conf/eccv/Shi2016Embedding}. In the two early works, Yi \etal \cite{yi2014deep} and Li \etal \cite{DBLP:conf/cvpr/LiZXW14} use image pairs and inject part priors into the learning process. In later works, Varior \etal \cite{DBLP:conf/eccv/VariorSLXW16} incorporate long short-term memory (LSTM) modules into a siamese network. LSTMs process image parts sequentially so that the spatial connections can be memorized
to enhance the discriminative ability of the deep features. Varior \etal \cite{DBLP:conf/eccv/VariorHW16}  insert a gating function after each convolutional layer to capture effective subtle patterns between image pairs. The above-mentioned methods are effective in learning image similarities in an adaptive manner, but may have efficiency problems under large-scale galleries.

The second type of CNN-based re-ID methods focuses on feature learning, which categorizes the training samples into pre-defined classes and the FC descriptor is used for retrieval \cite{DBLP:journals/corr/ZhengYH16,su2016deep,DBLP:conf/cvpr/XiaoLOW16}.  In \cite{DBLP:journals/corr/ZhengYH16,zheng2016person}, the classification CNN model is fine-tuned using either the video frames or image bounding boxes to learn a discriminative embedding for pedestrian retrieval.  Xiao \etal \cite{DBLP:conf/cvpr/XiaoLOW16} propose learning generic feature representations from multiple re-ID datasets jointly. To deal with spatial misalignment, Zheng \etal \cite{DBLP:journals/corr/ZhengHLY17} propose the PoseBox structure similar to the pictorial structure \cite{cheng2011custom} to learn pose invariant embeddings. To take advantage of both the feature learning and similarity learning, Zheng \etal \cite{DBLP:journals/corr/ZhengZY16} and Geng \etal \cite{geng2016deep}  combine the contrastive loss and the identification loss to improve the discriminative ability of the learned feature embedding, following the success in face verification \cite{sun2014deep}. This paper adopts the classification mode, which is shown to produce competitive accuracy without losing efficiency potentials.

\textbf{PCANet and truncated SVD for CNN.}
We clarify the difference between SVDNet and several ``look-alike'' works.
The PCANet \cite{DBLP:journals/tip/ChanJGLZM15} is proposed for image classification. It is featured by cascaded principal component analysis (PCA) filters. PCANet is related to SVDNet in that it also learns orthogonal projection directions to produce the filters. The proposed SVDNet  differs from PCANet in two major aspects. First, SVDNet performs SVD on the weight matrix of CNN, while PCANet performs PCA on the raw data and feature. Second, the filters in PCANet are learned in an unsupervised manner, which does not rely on back propagation as in the case of SVDNet. In fact, SVDNet manages a stronger connection between CNN and SVD. SVDNet's parameters are learned through back propagation and decorrelated iteratively using SVD.

Truncated SVD \cite{DBLP:conf/nips/DentonZBLF14,DBLP:conf/interspeech/XueLG13} is widely used for CNN model compression. SVDNet departs from it in two aspects. First, truncated SVD decomposes the weight matrix in FC layers and reconstructs it with several dominant singular vectors and values. SVDNet does not reconstruct the weight matrix but replaces it with an orthogonal matrix, which is the product of the left unitary matrix and the singular value matrix. Second, Truncated SVD reduces the model size and testing time at the cost of acceptable precision loss, while SVDNet significantly improves the retrieval accuracy without impact on the model size.   

\textbf{Orthogonality in the weight matrix.}
We note a concurrent work \cite{DBLP:conf/cvpr/Di17} which also aims to orthogonalize the CNN filters, yet our work is different from \cite{DBLP:conf/cvpr/Di17}. In  \cite{DBLP:conf/cvpr/Di17}, the regularization effect of orthogonalization benefits the back-propagation of very deep networks, thus improving the classification accuracy. The regularization proposed in \cite{DBLP:conf/cvpr/Di17} may not directly benefit the embedding learning process. But in this paper, orthogonalization is used to generate decorrelated descriptors suitable for retrieval. Our network may not be suitable for improving classification. 


\section{Proposed Method}
This section describes the structure of SVDNet, its training strategy, and its working mechanism.

\subsection{Architecture} \label{sec:architecture}
\begin{figure}[t]
\begin{center}
\includegraphics[width=1\linewidth]{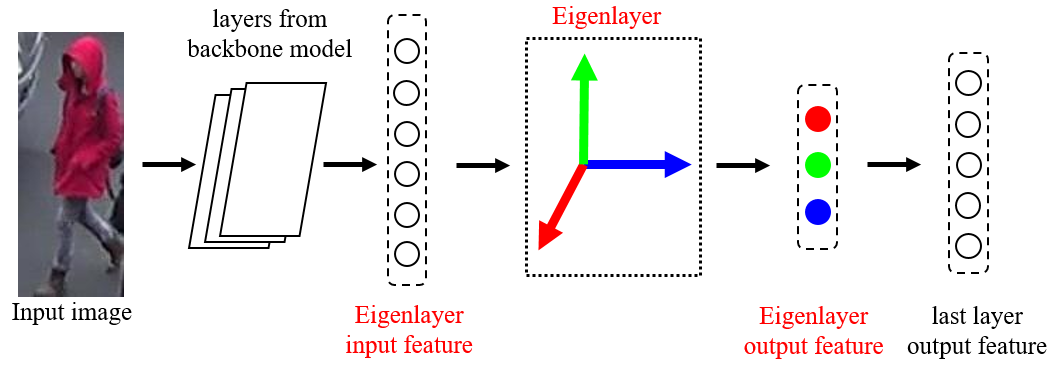}
\end{center}
\setlength{\abovecaptionskip}{0cm} 
   \caption{The architecture of SVDNet. It contains an Eigenlayer before the last FC layer of the backbone model. The weight vectors of the Eigenlayer are expected to be orthogonal. In testing, either the \emph{Eigenlayer input feature} or the \emph{Eigenlayer output feature} is employed for retrieval.} 
\label{fig:svdnet}
\end{figure}

SVDNet mostly follows the backbone networks, \eg, CaffeNet and ResNet-50. The only difference is that SVDNet uses the Eigenlayer as the second last FC layer, as shown in Fig. \ref{fig:svdnet}, the Eigenlayer contains an orthogonal weight matrix and is a linear layer without bias. The reason for not using bias is that the bias will disrupt the learned orthogonality. In fact, our preliminary experiments indicate that adding the ReLU activation and the bias term slightly compromises the re-ID performance, so we choose to implement the Eigenlayer based on a linear layer. The reason for positioning Eigenlayer at the second last FC layer, rather than the last one is that the model fails to converge when orthogonality is enforced on the last FC layer, which might be due to that the correlation of weight vectors in the last FC layer is determined by the training sample distribution, as explained in the introduction.
During training, the input feature from a previous layer is passed through the Eigenlayer. Its inner products with the weight vectors of the Eigenlayer form the output feature, which is fully connected to the last layer of $c$-dim, where $c$ denotes the number of training classes. 

During testing, we extract the learned embeddings for the query and gallery images. In this step, we can use either the input or the output of Eigenlayer for feature representation, as shown in Fig. \ref{fig:svdnet}. Our experiment shows that using the two features can achieve similar performance, indicating that the orthogonality of Eigenlayer improves the performance of not only output but also input. The reason is a bit implicit, and we believe it originates from the back-propagation training of CNN, during which the orthogonal characteristic of weight matrix within the Eigenlayer will directly impact the characteristic of its input feature.

\subsection{Training SVDNet} \label{sec:training}
The algorithm of training SVDNet is presented in Alg. \ref{alg:svd}. We first briefly introduce Step 0 and then describe the restraint and relaxation Iteration (RRI) (Step 1, 2, 3).

\textbf{Step 0.} We first add a linear layer to the network. Then the network is fine-tuned till convergence. Note that after Step 0, the weight vectors in the linear layer are still highly correlated. In the experiment, we will present the re-ID performance of the CNN model after Step 0. Various output dimensions of the linear layer will be evaluated.

\textbf{Restraint and Relaxation Iteration (RRI).} It is the key procedure in training SVDNet. Three steps are involved.
\begin{itemize}
\item Decorrelation. We perform SVD on the weight matrix as follows:
\begin{align}
W=USV^\mathrm{T},
\label{eq:SVD0}
\end{align}
where $W$ is the weight matrix of the linear layer, $U$ is the left-unitary matrix, $S$ is the singular value matrix, and $V$ is the right-unitary matrix.
After the decomposition, we replace $W$ with $US$. Then the linear layer uses all the eigenvectors of $WW^\mathrm{T}$ as weight vectors and is named as Eigenlayer. 
\item Restraint. The backbone model is fine-tuned till convergence, but the Eigenlayer is \emph{fixed}.
\item Relaxation. The fine-tuning goes on for some more epochs with Eigenlayer \emph{unfixed}. 
\end{itemize}
 
After Step 1 and Step 2, the weight vectors are orthogonal, \ie, in an eigen state. But after Step 3, \ie, relaxation training, $W$ shifts away from the eigen state. So the training procedure enters another iteration $t\;(t = 1,\ldots,T)$ of ``restraint and relaxation''. 

\begin{algorithm}[t]
\caption{Training SVDNet}
\textbf{Input:} a pre-trained CNN model, re-ID training data.\\
\textbf{0.} Add the Eigenlayer and fine-tune the network.\\
\For{$t\gets1\ to \ T$}
{ 
\textbf{1. Decorrelation:} Decompose $W$ with SVD decomposition, and then update it: {$W \gets US$}\\
\textbf{2. Restraint:} Fine-tune the network with the Eigenlayer fixed \\
\textbf{3. Relaxation:} Fine-tune the network with the Eigenlayer unfixed \\
} 
\textbf{Output:} a fine-tuned CNN model, \ie, SVDNet.\\
\label{alg:svd}
\end{algorithm}

Albeit simple, the mechanism behind the method is interesting. We will try to provide insight into the mechanism in Section \ref{sec:mechanism}. During all the analysis involved, CaffeNet pre-trained on ImageNet is chosen as the backbone. 

\subsection{Mechanism Study} \label{sec:mechanism}

\textbf{Why is SVD employed?} 
Our key idea is to find a set of orthogonal projection directions based on what CNN has already learned from training set. Basically, for a linear layer, a set of basis in the range space of $W$ (\ie, linear subspace spanned by column vectors of $W$) is a potential solution. In fact, there exists numerous sets of orthogonal basis. So we decide to use the singular vectors of $W$ as new projection directions and to weight the projection results with the corresponding singular values. That is, we replace \(W=USV^\mathrm{T}\) with \(US\). By doing this, {the discriminative ability  of feature representation over the whole sample space will be maintained}. We make a mathematical proof as follows:

Given two images \(x_i\) and \(x_j\), we denote $\vec{h_i}$ and $\vec{h_j}$ as the corresponding features  before the Eigenlayer, respectively. $\vec{f_i}$ and $\vec{f_j}$ are their output features from the Eigenlayer. The Euclidean distance \(D_{ij}\) between the features of \(x_i\) and \(x_j\) is calculated by:
{\setlength\abovedisplayskip{5pt}
\setlength\belowdisplayskip{7pt}
\begin{align}
D_{ij}&={\Vert\vv{f_i}-\vv{f_j}\Vert}_{2}=\sqrt{(\vv{f_i}-\vv{f_j})^\mathrm{T}(\vv{f_i}-\vv{f_j})}\nonumber\\
&=\sqrt{(\vv{h_i}-\vv{h_j})^\mathrm{T}WW^\mathrm{T}(\vv{h_i}-\vv{h_j})}\nonumber\\
&=\sqrt{(\vv{h_i}-\vv{h_j})^\mathrm{T}USV^\mathrm{T}VS^\mathrm{T}U^\mathrm{T}(\vv{h_i}-\vv{h_j})},
\label{eq:eu}
\end{align}
}where $U$, $S$ and $V$ are defined in Eq. \ref{eq:SVD0}. Since $V$ is a unit orthogonal matrix, Eq. \ref{eq:eu} is equal to:
\begin{equation}D_{ij}=\sqrt{(\vv{h_i}-\vv{h_j})^\mathrm{T}USS^\mathrm{T}U^\mathrm{T}(\vv{h_i}-\vv{h_j})}
\label{eq:svd}
\end{equation}
Eq. \ref{eq:svd} suggests that when changing $W=USV^\mathrm{T}$ to $US$, $D_{ij}$ remains unchanged. \textbf{Therefore, in Step 1 of Alg. \ref{alg:svd}, the discriminative ability (re-ID accuracy) of the fine-tuned CNN model is 100\% preserved.}

There are some other decorrelation methods in addition to SVD. But these methods do not preserve the discriminative ability of the CNN model. To illustrate this point, we compare SVD with several competitors below. 
\begin{enumerate}
\setlength{\itemsep}{0.4ex} %
\item Use the originally learned $W$ (denoted by $Orig$).
\item Replace $W$ with $US$ (denoted by $US$).
\item Replace $W$ with $U$ (denoted by $U$).
\item Replace $W$ with $UV^\mathrm{T}$ (denoted by $UV^\mathrm{T}$).
\item  Replace $W=QR$ (Q-R decomposition) with $QD$, where $D$ is the diagonal matrix extracted from the upper triangle matrix $R$ (denoted by $QD$).
\end{enumerate}

Comparisons on Market-1501 \cite{DBLP:conf/iccv/ZhengSTWWT15} are provided in Table \ref{table:USQR}. We replace the FC layer with a 1,024-dim linear layer and fine-tune the model till convergence (Step 0 in Alg. \ref{alg:svd}). We then replace the fine-tuned $W$ with methods 2 - 5. All the four decorrelation methods 2 - 5 update $W$ to be an orthogonal matrix, but Table \ref{table:USQR} indicates that only replacing $W$ with $US$ retains the re-ID accuracy,  while the others degrade the performance.

\textbf{When does performance improvement happen?} As proven above, Step 1 in Alg. \ref{alg:svd}, \ie, replacing $W=USV^\mathrm{T}$ with $US$, does not bring an immediate accuracy improvement, but keeps it unchanged. Nevertheless, after this operation, the model has been pulled away from the original fine-tuned solution, and the classification loss on the training set will increase by a certain extent. 
Therefore, Step 2 and Step 3 in Alg. \ref{alg:svd} aim to fix this problem. The major effect of these two steps is to improve the discriminative ability of the input feature as well as the output feature of the Eigenlayer (Fig. \ref{fig:svdnet}). 
On the one hand, the restraint step learns the upstream and downstream layers of the Eigenlayer, which still preserves the orthogonal property. We show in Fig. \ref{fig:RRI} that this step improves the accuracy. On the other hand, the relaxation step will make the model deviate from orthogonality again, but it reaches closer to convergence. This step, as shown in Fig. \ref{fig:RRI}, deteriorates the performance. But within an RRI, the overall performance improves. Interestingly, when educating children, an alternating rhythm of relaxation and restraint is also encouraged.

\begin{table}[!t]
\setlength{\tabcolsep}{5.9pt}
\renewcommand\arraystretch{1.1}
\begin{center}
\begin{tabular}{|l|c|c|c|c|c|}
\hline
Methods & $Orig$ & $US$ & $U$ & $UV^\mathrm{T}$ & $QD$ \\
\hline
rank-1 &63.6&63.6&61.7&61.7&61.6\\
\hline
mAP &39.0&39.0&37.1&37.1&37.3\\
\hline
\end{tabular}
\end{center}
\setlength{\abovecaptionskip}{-0cm} 
\caption{Comparison of decorrelation methods in Step 1 of Alg. \ref{alg:svd}. Market-1501 and CaffeNet are used. We replace FC7 with a 1,024-dim linear layer. Rank-1 (\%) and mAP (\%) are shown.}
\label{table:USQR}
\end{table}


\textbf{Correlation diagnosing.} Till now, we have not provided a metric how to evaluate vector correlations.  In fact, the correlation between two vectors can be estimated by the correlation coefficient. However, to the best of our knowledge, it lacks an evaluation protocol for diagnosing the \emph{overall} correlation of a vector set. In this paper, we propose to evaluate the overall correlation as below. Given a weight matrix $W$, we define the gram matrix of $W$ as,
\begin{align}
G=W^\mathrm{T}W
&=\begin{bmatrix}
\vv{w_1}^\mathrm{T}\vv{w_1} &\vv{w_1}^\mathrm{T}\vv{w_2} &\cdots &\vv{w_1}^\mathrm{T}\vv{w_k}\\
\\
\vv{w_2}^\mathrm{T}\vv{w_1} &\vv{w_2}^\mathrm{T}\vv{w_2} &\cdots &\vv{w_2}^\mathrm{T}\vv{w_k}\\
\\
\\
\vv{w_k}^\mathrm{T}\vv{w_1} &\vv{w_k}^\mathrm{T}\vv{w_2} &\cdots &\vv{w_k}^\mathrm{T}\vv{w_k}\end{bmatrix}\nonumber\\
&=\begin{bmatrix} 
\hspace{0.30cm} g_{11} & \hspace{0.50cm} g_{12} &\hspace{0.35cm}\cdots&\hspace{0.20cm} g_{1k} $\quad$ \\
\\
\hspace{0.30cm} g_{21} & \hspace{0.50cm} g_{22} &\hspace{0.35cm}\cdots&\hspace{0.20cm} g_{2k} $\quad$ \\
\\
\\
\hspace{0.30cm} g_{k1} & \hspace{0.50cm} g_{k2} &\hspace{0.35cm}\cdots&\hspace{0.20cm} g_{kk} $\quad$ 
\end{bmatrix},
\end{align}    
where $k$ is the number of weight vectors in $W$ ($k$ = 4,096 in FC7 of CaffeNet), $g_{ij}\,(i,j = 1,...,k)$ are the entries in $W$, and $w_i\,(i=1,...,k)$ are the weight vectors in $W$. Given $W$, we define  $S(\cdot)$ as a metric to denote the extent of correlation between all the column vectors of $W$: 
\begin{equation}\label{eq:sw}
S(W)=\frac{\sum_{i=1}^{k}g_{ii}}{\sum_{i=1}^{k}\sum_{j=1}^{k}\vert g_{ij}\vert}.
\end{equation}
From Eq. \ref{eq:sw}, we can see that the value of $S(W)$ falls within $[\frac{1}{k}, 1]$. $S(W)$ achieves the largest value $1$ only when $W$ is an orthogonal matrix, \ie, $g_{ij} = 0, \mbox{if } i\neq j$. $S(W)$ has the smallest value $\frac{1}{k}$ when all the weight vectors are totally the same, \ie, $g_{ij} = 1,  \forall i,j$. So when $S(W)$ is close to $1/k$ or is very small, the weight matrix has a high correlation extent. For example, in our baseline, when directly fine-tuning a CNN model (without SVDNet training) using CaffeNet, $S(W_{\mbox{FC7}})=0.0072$, indicating that the weight vectors in the FC7 layer are highly correlated. As we will show in Section \ref{sec:RRI}, $S$ is an effective indicator to the convergence of SVDNet training.

\textbf{Convergence Criteria for RRI.}
When to stop RRI is a non-trivial problem, especially in application. We employ Eq. \ref{eq:sw} to evaluate the orthogonality of $W$ after the relaxation step and find that $S(W)$ increases as the iteration goes on. It indicates that the correlation among the weight vectors in $W$ is reduced step-by-step with RRI. So when $S(W)$ becomes stable, the model converges, and RRI stops. Detailed observations can be accessed in Fig. \ref{fig:RRI}.

\section{Experiment}

\subsection{Datasets and Settings}
\textbf{Datasets.} 
This paper uses three datasets for evaluation, \ie,  \textbf{Market-1501} \cite{DBLP:conf/iccv/ZhengSTWWT15}, \textbf{CUHK03} \cite{DBLP:conf/cvpr/LiZXW14} and \textbf{DukeMTMC-reID} \cite{ristani2016MTMC,zheng2017unlabeled}. The Market-1501 dataset contains 1,501 identities, 19,732 gallery images and 12,936 training images captured by 6 cameras. All the bounding boxes are generated by the DPM detector \cite{felzenszwalb2008discriminatively}. Most experiments relevant to mechanism study are carried out on Market-1501. The CUHK03 dataset contains 13,164 images of 1,467 identities. Each identity is observed by 2 cameras. CUHK03 offers both hand-labeled and DPM-detected bounding boxes, and we use the latter in this paper. For CUHK03, 20 random train/test splits are performed, and the averaged results are reported. The DukeMTMC-reID dataset is collected with 8 cameras and used for cross-camera tracking. We adopt its re-ID version benchmarked in \cite{zheng2017unlabeled}. It contains 1,404 identities (one half for training, and the other for testing), 16,522 training images, 2,228 queries, and 17,661 gallery images. For Market-1501 and DukeMTMC-reID, we use the evaluation packages provided by \cite{DBLP:conf/iccv/ZhengSTWWT15} and \cite{zheng2017unlabeled}, respectively.

\begin{table*}[]
\begin{center}
\begin{tabular}{l|c|c c c c|c c c c|c c c c}
\hline
\multicolumn{1}{c|}{\multirow{2}{*}{Models \& Features}}&\multicolumn{1}{c|}{\multirow{2}{*}{dim}}&\multicolumn{4}{c|}{Market-1501} & \multicolumn{4}{c|}{CUHK03} & \multicolumn{4}{c}{DukeMTMC-reID}\\ 
\cline{3-14}
\multicolumn{1}{c|}{}&\multicolumn{1}{c|}{}&\multicolumn{1}{c}{R-1}&{R-5}&{R-10}&{mAP}&{R-1}&{R-5}&{R-10}&{mAP}&{R-1}&{R-5}&{R-10}&{mAP}\\
\hline
Baseline(C) FC6 & 4096 &55.3&75.8&81.9&30.4 & 38.6 & 66.4& 76.8&45.0 &46.9&63.2&69.2&28.3\\

Baseline(C) FC7&4096   &54.6&75.5&81.3&30.3  &42.2&70.2&80.4&48.6    &45.9&62.0&69.7&27.1\\

SVDNet(C) FC6 &4096  &\textbf{80.5}&\textbf{91.7}&\textbf{94.7}&\textbf{55.9}
  &\textbf{68.5}&\textbf{90.2}&\textbf{95.0}&\textbf{73.3}   &\textbf{67.6}&\textbf{80.5}&\textbf{85.7}&\textbf{45.8}\\

SVDNet(C) FC7  &1024 &79.0&91.3&94.2&54.6  &66.0&89.4&93.8&71.1    &66.7&80.5&85.1&44.4\\

\hline
Baseline(R) Pool5&2048   &73.8&87.6&91.3&47.9  &66.2&87.2&93.2&71.1    &65.5&78.5&82.5&44.1\\

Baseline(R) FC   &$N$      &71.1&85.0&90.0&46.0  &64.6&89.4&95.0&70.0    &60.6&76.0&80.9&40.4\\

SVDNet(R) Pool5 &2048  &\textbf{82.3}&\textbf{92.3}&\textbf{95.2}&\textbf{62.1}  
&\textbf{81.8}&\textbf{95.2}&97.2&\textbf{84.8}   
 &\textbf{76.7}&\textbf{86.4}&\textbf{89.9}&\textbf{56.8}\\

SVDNet(R) FC    &1024  &81.4&91.9&94.5&61.2  &81.2&95.2&\textbf{98.2}&84.5    &75.9&86.4&89.5&56.3\\
\hline 
\end{tabular}

\end{center}
\setlength{\abovecaptionskip}{-0cm} 
\caption{Comparison of the proposed method with baselines. C: CaffeNet. R: ResNet-50. In ResNet Baseline, ``FC'' denotes the last FC layer, and its output dimension $N$ changes with the number of training identities, \ie, 751 on Market-1501, 1,160 on CUHK03 and 702 on DukeMTMC-reID. For SVDNet based on ResNet, the Eigenlayer is denoted by ``FC'', and its output dimension is set to 1,024.}
\label{table:cmpbasl}
\end{table*}

For performance evaluation on all the 3 datasets, we use both the Cumulative Matching Characteristics (CMC) curve and the mean Average Precision (mAP).

\textbf{Backbones.} We mainly use two networks pre-trained on ImageNet \cite{DBLP:conf/cvpr/DengDSLL009} as backbones, \ie, CaffeNet \cite{DBLP:conf/nips/KrizhevskySH12} and ResNet-50 \cite{DBLP:conf/cvpr/HeZRS16}. 
When using CaffeNet as the backbone, we directly replace the original FC7 layer with the Eigenlayer, in case that one might argue that the performance gain is brought by deeper architecture. When using ResNet-50 as the backbone, we have to insert the Eigenlayer before the last FC layer because ResNet has no hidden FC layer and the influence of adding a layer into a 50-layer architecture can be neglected.
In several experiments on Market-1501, we additionally use  VGGNet \cite{simonyan2014very} and a Tiny CaffeNet as backbones to demonstrate the effectiveness of SVDNet on different architectures. The Tiny CaffeNet is generated by reducing the FC6 and FC7 layers of CaffeNet to containing 1024 and 512 dimensions, respectively.

\subsection{Implementation Details}\label{sec:details}
\textbf{Baseline.} Following the practice in \cite{DBLP:journals/corr/ZhengYH16}, baselines using CaffeNet and ResNet-50 are fine-tuned with the default parameter settings except that the output dimension of the last FC layer is set to the number of training identities. The CaffeNet Baseline is trained for 60 epochs with a learning rate of 0.001 and then for another 20 epochs with a learning rate of 0.0001. The ResNet Baseline is trained for 60 epochs with learning rate initialized at 0.001 and reduced by 10 on 25 and 50 epochs. During testing, the FC6 or FC7 descriptor of CaffeNet and the Pool5 or FC descriptor of ResNet-50 are used for feature representation.

On Market-1501, CaffeNet and Resnet-50 achieves rank-1 accuracy of 55.3\% (73.8\%) with the FC6 (Pool5) descriptor, which is consistent with the results in \cite{DBLP:journals/corr/ZhengYH16}.

\textbf{Detailed settings.} 
CaffeNet-backboned SVDNet takes 25 RRIs to reach final convergence. For both the restraint stage and the relaxation stage within each RRI except the last one, we use 2000 iterations and fix the learning rate at 0.001. For the last restraint training, we use 5000 iterations (learning rate 0.001) + 3000 iterations (learning rate 0.0001). The batch size is set to 64. ResNet-backboned SVDNet takes 7 RRIs to reach final convergence. For both the restraint stage and the relaxation stage within each RRI, we use 8000 iterations and divide the learning rate by 10 after 5000 iterations. The initial learning rate for the 1st to the 3rd RRI is set to 0.001, and the initial learning rate for the rest RRIs is set to 0.0001. The batch size is set to 32.

The output dimension of Eigenlayer is set to be 1024 in all models, yet the influence of this hyper-parameter is to be analyzed in Section \ref{sec:dimension}. The reason of using different times of RRIs for different backbones is to be illustrated in Section \ref{sec:RRI}.

\subsection{Performance Evaluation}\label{sec:evaluation}
\textbf{The effectiveness of SVDNet.} We comprehensively evaluate the proposed SVDNet on all the three re-ID benchmarks. The overall results are shown in Table \ref{table:cmpbasl}.

\begin{figure}[t]
\begin{center}
\includegraphics[width=1\linewidth]{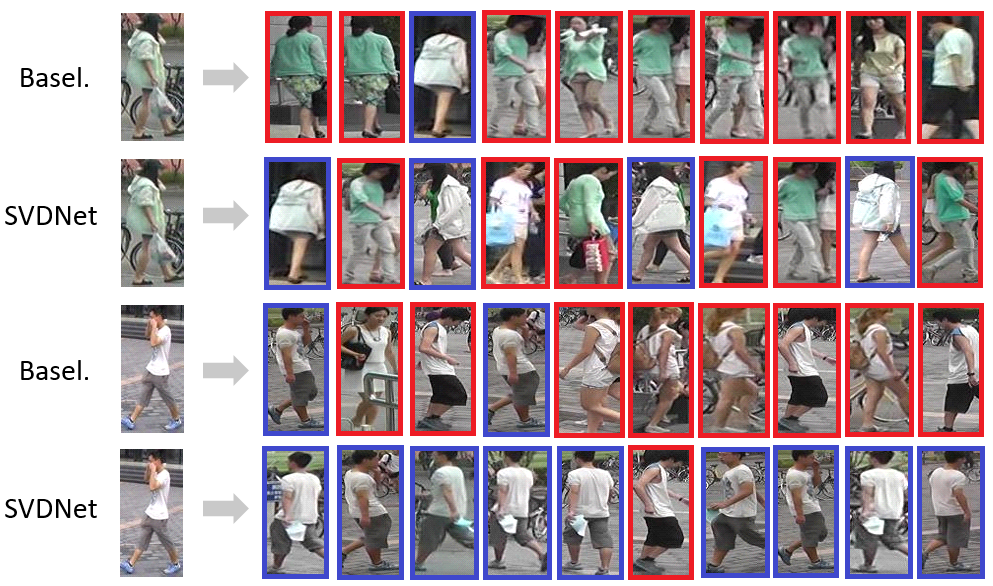}
\end{center}
\setlength{\abovecaptionskip}{-0cm} 
   \caption{Sample retrieval results on Market-1501. In each row, images are arranged in descending order according to their similarities with the query on the left. The true and false matches are in the blue and red boxes, respectively.} 
\label{fig:examples}
\end{figure}

\setlength{\tabcolsep}{5pt}
\begin{table}[t]
\begin{center}
\begin{tabular}{l|cc|cc}
\hline
\multicolumn{1}{l|}{\multirow{2}{*}{Methods}}&\multicolumn{2}{c|}{Market-1501}&\multicolumn{2}{c}{CUHK03}\\
\cline{2-5}
\multicolumn{1}{c|}{}&rank-1&mAP&rank-1&mAP\\
\hline
\hline
LOMO+XQDA\cite{DBLP:conf/cvpr/LiaoHZL15}     &43.8&22.2      &44.6&51.5\\
CAN\cite{DBLP:journals/corr/LiuFQJY16}           &48.2&24.4      &63.1&-   \\
SCSP\cite{DBLP:conf/cvpr/ChenYCZ16}          &51.9&26.4      &-   &-   \\
Null Space\cite{DBLP:conf/cvpr/ZhangXG16}    &55.4&29.9      &54.7&-   \\
DNS\cite{DBLP:conf/cvpr/ZhangXG16}           &61.0&35.6      &54.7&-   \\
LSTM Siamese\cite{DBLP:conf/eccv/VariorSLXW16}  &61.6&35.3      &57.3&46.3\\
MLAPG\cite{DBLP:conf/iccv/LiaoL15}         &-   &-         &58.0&-   \\
Gated SCNN\cite{DBLP:conf/eccv/VariorHW16}    &65.9&39.6      &61.8&51.3\\
ReRank (C) \cite{DBLP:conf/cvpr/ZhongZCL17}     &61.3&46.8      &58.5&64.7\\
ReRank (R) \cite{DBLP:conf/cvpr/ZhongZCL17}     &77.1&\textcolor{red}{63.6}      &64.0&69.3\\
\hline
PIE (A)* \cite{DBLP:journals/corr/ZhengHLY17}        &65.7&41.1      &62.6&67.9\\
PIE (R)* \cite{DBLP:journals/corr/ZhengHLY17}        &79.3&56.0      &67.1&71.3\\
SOMAnet (VGG)* \cite{barbosa2017looking} &73.9&47.9&72.4&-\\
DLCE (C)* \cite{DBLP:journals/corr/ZhengZY16}       &62.1&39.6      &59.8&65.8\\
DLCE (R)* \cite{DBLP:journals/corr/ZhengZY16}       &79.5&59.9      &\textcolor{red}{83.4}&\textcolor{blue}{86.4}\\
Transfer (G)* \cite{geng2016deep}  &\textcolor{blue}{83.7}&\textcolor{blue}{65.5}      &\textcolor{blue}{84.1}&-\\
\hline
SVDNet(C)     &\textcolor{green}{80.5}&55.9      &68.5&\textcolor{green}{73.3}\\
SVDNet(R,1024-dim)&\textcolor{red}{82.3}&\textcolor{green}{62.1}     &\textcolor{green}{81.8}&\textcolor{red}{84.8}\\
\hline
\end{tabular}
\end{center}
\setlength{\abovecaptionskip}{-0cm} 
\caption{Comparison with state of the art on Market-1501 (single query) and CUHK03. * denotes unpublished papers. Base networks are annotated. C: CaffeNet, R: ResNet-50, A: AlexNet, G: GoogleNet \cite{szegedy2015going}. The best, second and third highest results are in \textcolor{blue}{blue}, \textcolor{red}{red} and \textcolor{green}{green}, respectively.}
\label{table:sota-market}
\end{table}

\begin{table}
\begin{center}
\begin{tabular}{l|cc|cc}
\hline
\multicolumn{1}{l|}{\multirow{2}{*}{Methods}}&\multicolumn{2}{c|}{DukeMTMC-reID}&\multicolumn{2}{c}{CUHK03-NP}\\
\cline{2-5}
\multicolumn{1}{c|}{}&rank-1&mAP&rank-1&mAP\\
\hline
BoW+kissme \cite{DBLP:conf/iccv/ZhengSTWWT15} & 25.1 & 12.2 & 6.4 & 6.4\\
LOMO+XQDA \cite{DBLP:conf/cvpr/LiaoHZL15} & 30.8 & 17.0 & 12.8 & 11.5 \\
Baseline (R) & 65.5 & 44.1 & 21.3 & 19.7 \\ 
GAN (R) \cite{zheng2017unlabeled} & {67.7}  & {47.1} & - & - \\
PAN (R) \cite{DBLP:journals/corr/PAN} & 71.6& 51.5 & 36.3 &34.0\\
\hline
SVDNet (C) &67.6&45.8 &27.7 & 24.9 \\
SVDNet (R) &\textbf{76.7}& \textbf{56.8} & \textbf{41.5} & \textbf{37.3} \\
\hline
\end{tabular}
\end{center}
\setlength{\abovecaptionskip}{0cm} 
\setlength{\belowcaptionskip}{0pt} 
\caption{Comparison with the state of the art on DukeMTMC-reID and CUHK03-NP. Rank-1 accuracy (\%) and mAP (\%) are shown. For fair comparison, all the results are maintained without post-processing methods.}
\label{table:duke}
\end{table}

\begin{figure*}[t]
\begin{minipage}[t]{0.48\linewidth}
\centering
	\includegraphics[width=1\linewidth]{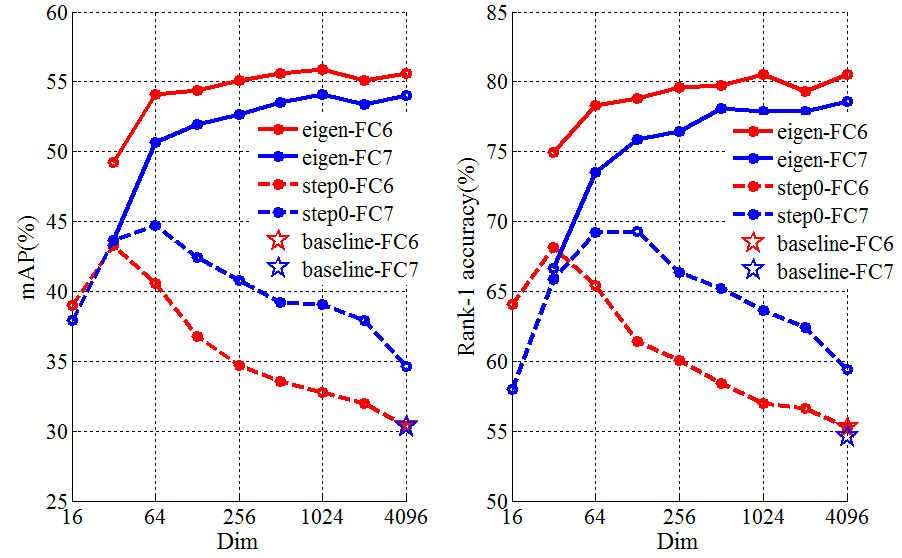}
\caption*{(a) CaffeNet-backboned SVDNet}
\end{minipage}
\hfill
\begin{minipage}[t]{0.48\linewidth}
\centering
	\includegraphics[width=0.985\linewidth]{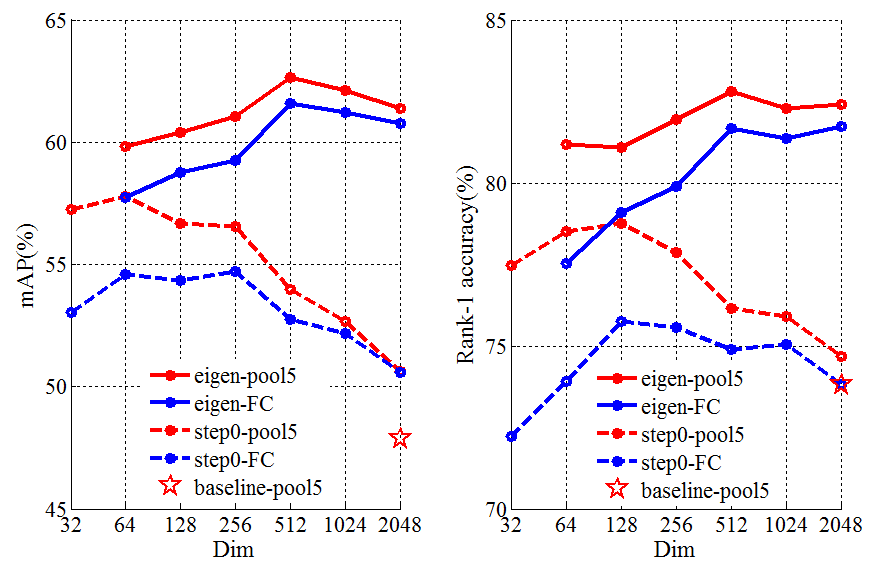}
\caption*{(b) ResNet-backboned SVDNet}
\end{minipage}
\setlength{\abovecaptionskip}{0cm}
\caption{Dimension comparison on (a) CaffeNet-backboned and (b) ResNet-backboned. The marker prefixed by ``step0'' denotes that the corresponding model is trained without any RRI. The marker prefixed by ``eigen'' denotes that the corresponding model is trained with sufficient RRIs to final convergence. For (a), the output dimension of Eigenlayer is set to 16, 32, 64, 128, 256, 512, 1024, 2048 and 4096. For (b), the output dimension of Eigenlayer is set to 32, 64, 128, 256, 512, 1024 and 2048.}
\label{fig:dim}
\end{figure*}


The improvements achieved on both backbones are significant: When using CaffeNet as the backbone, the Rank-1 accuracy on Market-1501 rises from 55.3\% to 80.5\%, and the mAP rises from 30.4\% to 55.9\%. On CUHK03 (DukeMTMC-reID) dataset, the Rank-1 accuracy rises by +26.3\% (+20.7\%), and the mAP rises by +24.7\% (+17.5\%). When using ResNet as the backbone, the Rank-1 accuracy rises by +8.4\%, +15.6\% and +11.2\% respectively on Market-1501, CUHK03 and DukeMTMC-reID dataset. The mAP rises by +14.2\%, +13.7\% and +12.7\% correspondingly. Some retrieval examples on Market-1501 are shown in Fig. \ref{fig:examples}.


\textbf{Comparison with state of the art.}
We compare SVDNet with the state-of-the-art methods. Comparisons on Market-1501 and CUHK03 are shown in Table \ref{table:sota-market}. Comparing with already published papers, SVDNet achieves competitive performance. We report \textbf{rank-1 = 82.3\%, mAP = 62.1\% on Market-1501, and rank-1 = 81.8\%, mAP = 84.8\% on CUHK03}. The re-ranking method \cite{DBLP:conf/cvpr/ZhongZCL17} is higher than ours in mAP on Market-1501, because re-ranking exploits the relationship among the gallery images and results in a high recall. We speculate that this re-ranking method will also bring improvement for SVDNet. Comparing with the unpublished Arxiv papers, (some of) our numbers are slightly lower than \cite{geng2016deep} and \cite{DBLP:journals/corr/ZhengZY16}. Both works \cite{geng2016deep} and \cite{DBLP:journals/corr/ZhengZY16} combine the verification and classification losses, and we will investigate into integrating this strategy into SVDNet.

Moreover, the performance of SVDNet based on relatively simple CNN architecture is impressive. On Market-1501, CaffeNet-backboned SVDNet achieves 80.5\% rank-1 accuracy and 55.9\% mAP, exceeding other CaffeNet-based methods by a large margin. Additionally, using VGGNet and Tiny CaffeNet as backbone achieves 79.7\% and 77.4\% rank-1 accuracy respectively. On CUHK03, CaffeNet-backboned SVDNet even exceeds some ResNet-based competing methods except DLCE(R). This observation suggests that our method can achieve acceptable performance with high computing effectiveness. 

\begin{figure*}[t]
\begin{center}
\includegraphics[width=0.9\linewidth]{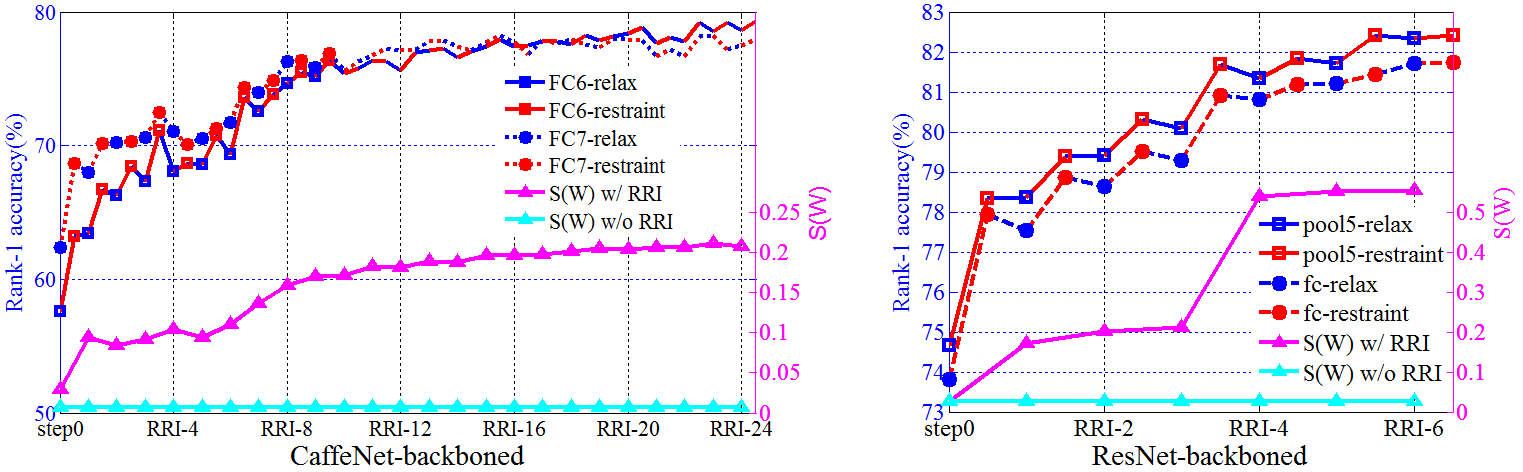}
\end{center}
\setlength{\abovecaptionskip}{0cm}
   \caption{Rank-1 accuracy and $S(W)$ (Eq. \ref{eq:sw}) of each intermediate model during RRI. Numbers on the horizontal axis denote the end of each RRI. SVDNet based on CaffeNet and ResNet-50 take about 25 and 7 RRIs to converge, respectively. Results before the $11th$ RRI is marked. $S(W)$ of models trained without RRI is also plotted for comparison.}
\label{fig:RRI}
\end{figure*}

In Table \ref{table:duke}, comparisons on DukeMTMC-reID and CUHK03 under a new training/testing protocol (denoted as CUHK03-NP) raised by \cite{DBLP:conf/cvpr/ZhongZCL17} are summarized. Relatively fewer results are reported because both DukeMTMC-reID and CUHK03-NP have only been recently benchmarked. On DukeMTMC-reID, this paper reports \textbf{rank-1 = 76.7\%, mAP = 56.8\%}, which is higher than the several competing methods including a recent GAN approach \cite{zheng2017unlabeled}. On CUHK03-NP, this paper reports \textbf{rank-1 = 41.5\%, mAP = 37.3\%}, which is also the highest among all the methods. 

\subsection{Impact of Output Dimension} \label{sec:dimension}
We vary the dimension of the output of Eigenlayer. Results of CaffeNet and ResNet-50 are drawn in Fig. \ref{fig:dim}.
 
When trained without RRI, the model has no intrinsic difference with a baseline model. It can be observed that the output dimension of the penultimate layer significantly influences the performance. As the output dimension increases, the re-ID performance first increases, reaches a peak and then drops quickly. In this scenario, we find that lowering the dimension is usually beneficial, probably due to the reduced redundancy in filters of FC layer. 

The influence of the output dimension on the final performance of SVDNet presents another trend. As the output dimension increases, the performance gradually increases until reaching a stable level, which suggests that our method is immune to harmful redundancy.

\subsection{RRI Boosting Procedure} \label{sec:RRI}
This experiment reveals how the re-ID performance changes after each restraint step and each relaxation step, and how SVDNet reaches the stable performance step by step. In our experiment, we use 25 epochs for both the restraint phase and the relaxation phase in one RRI. The output dimension of Eigenlayer is set to 2,048. Exhaustively, we test re-ID performance and $S(W)$ values of all the intermediate CNN models. We also increase the training epochs of baseline models to be equivalent of training SVDNet, to compare $S(W)$ of models trained with and without RRI. Results are shown in Fig. \ref{fig:RRI}, from which four conclusions can be drawn. 
\setlength{\tabcolsep}{8.3pt}
\begin{table}
\renewcommand\arraystretch{1.1}
\begin{center}
\begin{tabular}{l|ccccc}
\hline
Methods &$Orig$& $US$ & $U$ & $UV^\mathrm{T}$ & $QD$ \\
\hline
FC6(C) & 57.0 & 80.5 & 76.2 & 57.4 & 58.8\\
FC7(C) & 63.6 & 79.0 & 75.8 & 62.7 & 63.2\\
\hline
Pool5(R)&75.9 & 82.3 & 80.9 & 76.5 & 77.9\\
FC(R)   &75.1 & 81.4 &80.2  &74.8 & 77.3\\
\hline
\end{tabular}
\end{center}
\setlength{\abovecaptionskip}{0cm}
\caption{Comparison of the decorrelation methods specified in Section \ref{sec:mechanism}. Rank-1 accuracy (\%) on Market-1501 is shown. Dimension of output feature of Eigenlayer is set to 1024. We run sufficient RRIs for each method. }
\label{table:dec}
\end{table}

First, within each RRI, rank-1 accuracy takes on a pattern of ``increase and decrease'' echoing the restraint and relaxation steps: When $W$ is fixed to maintain orthogonality during restraint training, the performance increases, implying a boosting in the discriminative ability of the learned feature. Then during relaxation training, $W$ is unfixed, and the performance stagnates or even decreases slightly. Second, as the RRI goes, the overall accuracy increases, and reaches a stable level when the model converges. Third, it is reliable to use \(S(W)\) -- the degree of orthogonality --  as the convergence criteria for RRI. During RRI training, $S(W)$ gradually increases until reaching stability, while without RRI training, $S(W)$ fluctuates slightly around a relatively low value, indicating high correlation among weight vectors. Fourth, ResNet-backboned SVDNet needs much fewer RRIs to converge than CaffeNet-backboned SVDNet. 

\subsection{Comparison of Decorrelation Methods} In Section \ref{sec:mechanism}, several decorrelation methods are introduced. 
We show that only the proposed method of replacing $W$ with $US$ maintains the discriminative ability of the output feature of Eigenlayer, while all the other three methods lead to performance degradation to some extent. Here, we report their final performance when RRI training is used.

Results on Market-1501 are shown in Table \ref{table:dec}. It can be observed that the proposed decorrelating method, \ie, replacing $W$ with $US$, achieves the highest performance, followed by the ``$U$'', ``$QD$'' and ``$UV^\mathrm{T}$'' methods. In fact, the ``$UV^\mathrm{T}$'' method  does not bring about  observable improvement compared with ``$Orig$''. This experiment demonstrates that not only the orthogonality itself, but also the decorrelation approach, are vital for SVDNet.

\section{Conclusions}
In this paper, SVDNet is proposed for representation learning in pedestrian retrieval, or re-identification. Decorrelation is enforced among the projection vectors in the weight matrix of the FC layer. Through iterations of ``restraint and relaxation'', the extent of vector correlation is gradually reduced. In this process, the re-ID performance undergoes iterative ``increase and decrease'', and finally reaches a stable accuracy. Due to elimination of correlation of the weight vectors, the learned embedding better suits the retrieval task under the Euclidean distance. Significant performance improvement is achieved on the Market-1501, CUHK03, and DukeMTMC-reID datasets, and the re-ID accuracy is competitive with the state of the art.

In the future study, we will investigate more extensions of SVDNet to find out more about its working mechanism. We will also apply SVDNet on the generic instance retrieval problem.

{\small
\bibliographystyle{ieee}
\bibliography{egbib}
}

\end{document}